\documentclass[10pt,twocolumn,letterpaper]{article}

\usepackage[pagenumbers]{wacv}              

\usepackage{amsmath,amssymb}
\usepackage{stmaryrd}
\usepackage[accsupp]{axessibility}

\definecolor{wacvblue}{rgb}{0.21,0.49,0.74}
\usepackage[pagebackref,breaklinks,colorlinks,allcolors=wacvblue]{hyperref}

\usepackage{amsmath}
\usepackage{gensymb}
\usepackage{subcaption}
\usepackage{multirow}
\usepackage{booktabs}
\usepackage{makecell}
\usepackage{float}

\def\wacvPaperID{2684} 
\def\confName{WACV}
\def\confYear{2027}

\title{Spatiotemporal Reconstruction of Auroral Precipitations with Neural Fields}

\author{
Nazar Misyats$^{1}$ \qquad Mizuki Fukizawa$^{2}$ \qquad Yoshimasa Tanaka$^{2}$ \qquad Satoshi Ikehata$^{3}$\\
$^{1}$University of British Columbia, Canada
\qquad $^{2}$National Institute of Polar Research, Japan\\
$^{3}$National Institute of Informatics, Japan
\\
{\tt\small nmisyats@ece.ubc.ca, \{fukizawa.mizuki, ytanaka\}@nipr.ac.jp, sikehata@nii.ac.jp}
}

\newcommand{\R}{\mathbb{R}}
\newcommand{\dif}{\mathrm{d}}
\newcommand{\tmin}{\mathrm{min}}
\newcommand{\tmax}{\mathrm{max}}
\newcommand{\mr}[1]{\mathrm{#1}}

\newcommand{\intrange}[2]{\llbracket #1, #2 \rrbracket}

\newcommand{\volume}{\mathcal{V}}
\newcommand{\surface}{\mathcal{S}}
\newcommand{\erange}{\mathcal{E}}
\newcommand{\pos}{\mathbf{p}}
\newcommand{\postop}{\mathbf{x}}
\newcommand{\ro}{\mathbf{o}}
\newcommand{\rd}{\mathbf{d}}

\newcommand{\flux}{\Phi}
\newcommand{\eflux}{Q}
\newcommand{\emean}{\bar{E}}
\newcommand{\kernel}{K}

\newcommand{\volrate}{L}
\newcommand{\pix}{I}

\newcommand{\nbasis}{{n}}

\newcommand{\embed}{\mathbf{z}}
\newcommand{\innov}{\mathbf{u}}
\newcommand{\loss}{\mathcal{L}}

\begin{document}

\AddToHookNext{shipout/foreground}{%
  \put(0,0){%
    \makebox[\paperwidth][c]{%
      \raisebox{-0.45in}[0pt][0pt]{%
        \parbox{\textwidth}{\centering\scriptsize
          \copyright~2027 IEEE.  Personal use of this material is permitted.  Permission from IEEE must be obtained for all other uses, in any current or future media, including reprinting/republishing this material for advertising or promotional purposes, creating new collective works, for resale or redistribution to servers or lists, or reuse of any copyrighted component of this work in other works.}}}}}

\maketitle

\begin{abstract}
Auroras provide remote optical observations of energetic electrons precipitating into the upper atmosphere, and whose energy distribution encodes key information about large scale magnetospheric processes. Recovering the underlying electron precipitation field from ground-based camera images of auroras remains a challenging inverse problem. Existing auroral tomography techniques are often limited to simplified static settings, discrete spatial and energy grids, or strong assumptions on the electron energy spectrum. We propose to model the dynamic auroral electron precipitation field via a temporally conditioned neural network. The state of the precipitation field at each frame is represented by a per-frame latent code driven by learned discrete dynamics. A shared decoder maps horizontal positions and latent states to electron energy spectra. The predicted flux is converted to multi-wavelength volume emission rates and rendered into camera images accounting for delayed emissions. We evaluate the method on synthetic dynamic auroral sequences and real multi-view auroral videos, showing that the proposed representation produces temporally coherent precipitation reconstructions and image projections consistent with the observations.
\end{abstract}

\section{Introduction}

\begin{figure}
  \centering
  \includegraphics[width=\linewidth]{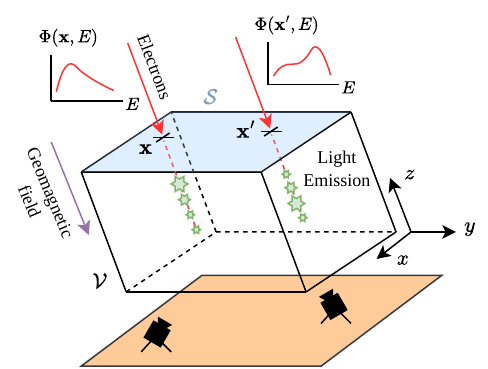}
  \caption{
    Visualization of auroral emission process. Electrons enter a volume \(\volume\) of the atmosphere through its top boundary \(\surface\). Their interaction with the atmospheric constituents inside \(\volume\) produces light emissions which are observed by ground-based cameras. The precipitation field \(\flux(\postop, E)\) describes the energy distribution of electrons entering through each position \(\postop\in\surface\).
    }
  \label{fig:volume_visualization}
\end{figure}

Auroras are optical emissions produced when charged particles, primarily electrons, precipitate along Earth's magnetic field lines and interact with the upper atmosphere \cite{reesPhysicsChemistryUpper1989}.
As illustrated in \cref{fig:volume_visualization}, electrons enter the atmosphere in a range of kinetic energies whose distribution -- energy spectra -- is directly related to the mechanisms driving magnetosphere-ionosphere coupling and important properties of ionospheric electrodynamics \cite{lyonsMagnetosphericProcessesLeading1997, robinsonCalculatingIonosphericConductances1987}.
Accurately determining the spatial structure and energy distribution of precipitating auroral electrons is therefore an important problem in space physics, with a long history of dedicated observation instruments and reconstruction methods \cite{stormerTwentyfiveYearsWork1930, mcilwainDirectMeasurementParticles1960, frankObservationsChargedParticle1971, freyOpticalTomographyAurora1998, redmonNewDMSPDatabase2017}.

Distributed camera networks provide long-duration, multi-wavelength and multi-view image sequences of auroras over a wide region of the sky.
Examples include ALIS \cite{brandstromAuroralLargeImaging2003}, MIRACLE \cite{sangalliPerformanceStudyNew2011} and WMI \cite{ogawaDevelopmentLowcostMultiwavelength2020} in Northern Scandinavia, and THEMIS GBO \cite{mendeTHEMISArrayGroundbased2008} in Canada and Alaska.
These networks record the continuous spatial and temporal evolution of auroral emissions at a much higher resolution and reliable rate than in-situ measurements from radars and satellites.
However, recovering the underlying precipitation field from ground-based observations is a severely ill-posed inverse problem. Cameras capture the line-of-sight integrals of the observable 3D emissions, which are only an indirect energy-dependent response to the 2D field of electron spectra entering the atmosphere.
The problem is further complicated by sparse viewpoints (five or less), partial camera calibration, sensor noise, and long exposure times.

Aurora computed tomography (ACT) reconstructs the 3D atmospheric emission field by performing a tomographic inversion of the multi-view observations \cite{freyAuroralEmissionProfiles1996, freyOpticalTomographyAurora1998, asoAnalysisAuroraStereo1990, nygrenApplicationStochasticInversion1996, enellTomographylikeRetrievalAuroral2012}. Physics-informed extensions recover the precipitating field jointly with the 3D emissions \cite{janhunenReconstructionElectronPrecipitation2001, partamiesTestingInversionMethod2004, hirschReconstructionFineScaleAuroral2016, fukizawaReconstructionPrecipitatingElectrons2022}.
Time-resolved reconstruction has also been performed by repeating the tomographic inversion on successive images or by combining high-speed optical observations with radar measurements \cite{simonwedlundEstimatingEnergySpectra2013,tuttleTemporalSpatialEvolution2014,hirschReconstructionFineScaleAuroral2016,fukizawaReconstructionPrecipitatingElectrons2022}, demonstrating the scientific value of dynamic reconstruction. They however generally consist in solving a large discretized inverse problem for each frame, and so do not exploit the temporal redundancy across long multi-view videos. ACT is also computationally expensive and sensitive to regularization and calibration choices.
Color-ratio methods avoid volumetric tomography, but require assuming a specific distribution family \cite{reesAuroralElectronEnergy1974,onoDerivationEnergyParameters1993}.

A recent work \cite{misyatsFastScalableReconstruction2026} represents the precipitating flux of electrons with a coordinate-based neural network that maps horizontal positions to local energy histograms and is optimized through a differentiable atmospheric and image formation model. This continuous representation substantially improves the efficiency and accuracy of static reconstruction over ACT, but retains the same steady-state formulation for individual images. This assumption is problematic for videos because some auroral emissions persist at time scales comparable to camera exposures. The brightness in one frame can therefore depend on precipitation in the preceding frames \cite{nozawaNewFivewavelengthPhotometer2018}. Independent static reconstructions both discard temporal redundancy and can attribute delayed emissions to the wrong precipitation state.

In this paper, we address these limitations with a physics-constrained neural field that jointly reconstructs a sequence of spatially and temporally varying precipitating electron spectra from sparse multi-view, multi-wavelength ground-based videos of auroras. Each frame is represented by a latent code generated by a recurrent process. A decoder then maps positions and frame codes to a continuous energy spectrum. A differentiable volume renderer converts the spectra to 3D emissions and camera observations. Our contributions are threefold: (1) to our knowledge, the first sequence-level neural-field formulation for reconstructing time-varying auroral electron precipitation; (2) a single optimization pipeline which couples neighboring frames and jointly estimates relative camera sensitivities; and (3) an evaluation on accurately simulated dynamic auroras with known precipitation fields and real multi-view recordings.
\section{Related work}

\subsection{Auroral Tomography}

Techniques for inverting auroral emissions have been an active field of study since the deployment of scientific auroral imaging systems.
Aurora computed tomography reconstructs the 3D emissive field in a discretized volume of the atmosphere from multi-view observations using tomographic inversion \cite{freyAuroralEmissionProfiles1996, freyOpticalTomographyAurora1998, asoAnalysisAuroraStereo1990, nygrenApplicationStochasticInversion1996, enellTomographylikeRetrievalAuroral2012}.
The energy spectra of precipitating electrons at the top of the volume can then be recovered from these 3D emissions by inverting a ionospheric emission model \cite{janhunenReconstructionElectronPrecipitation2001, partamiesTestingInversionMethod2004, hirschReconstructionFineScaleAuroral2016, fukizawaReconstructionPrecipitatingElectrons2022}.
Generalized ACT (G-ACT) combines both optical and radar measurements and recover the precipitating flux jointly with the tomographic reconstruction \cite{tanakaFeasibilityStudyGeneralizedAurora2011, fukizawaThreedimensionalIonosphericConductivity2023,
tanakaApplicationGeneralizedAurora2024}.
An alternative approach \cite{robertReconstructionElectronPrecipitation2023} instead fits the distribution that minimizes the discrepancies between the tomographic reconstruction and emissions obtained from numerical simulations of electron transport.

The applicability of ACT-based methods to complete video sequences remains however limited. First, ACT solves a large linear inverse problem with iterative optimization, which takes on the order of an hour per frame and does not scale to multi-hour videos. Second, due to the limited number of viewpoints and large number of degrees of freedom, the convergence of the reconstruction is highly sensitive to the choice of hyper-parameters, regularization coefficients, and knowledge of camera calibration, which requires repeated manual tuning.

\subsection{Coordinate-based Neural Fields}

Coordinate-based neural networks have gained a lot of attention in computer vision and graphics for solving continuous 3D inverse problems \cite{xieNeuralFieldsVisual2022}.
Rather than storing a scalar field over a discrete spatial grid, it is instead implicitly represented by a multi-layer perceptron (MLP) mapping continuous spatial coordinates to field values. The weights of the MLP are optimized to fit observations through a differentiable forward observation model.
Neural radiance fields (NeRFs) \cite{mildenhallNeRFRepresentingScenes2021} exploit this representation for the reconstruction of realistic 3D scenes as continuous density and radiance fields. Similar formulations have been successfully applied to tomographic and physics-based inverse problems in medical imaging \cite{arratiaNeuralFieldsHighly, sunCoILCoordinateBasedInternal2021}, astronomy \cite{levisGravitationallyLensedBlack2022, zhaoSingleViewRefractive2024, levisRevealingFineStructure2025} and microscopy \cite{zhongCryoDRGNReconstructionHeterogeneous2021}.

A recent work \cite{misyatsFastScalableReconstruction2026} proposed to model the auroral electron flux as a neural field. An MLP maps a continuous horizontal position at the top of the atmosphere to the histogram of the energy distribution. The MLP's parameters are optimized to reproduce the auroral emissions observed by multi-view cameras via differentiable rendering. Contrary to most neural field-based applications, where the modeled and learned quantity is the 3D field responsible for the observations, such as radiance or density, in auroral flux reconstruction, the observed emissions are an intermediate quantity physically related to the unobserved field being learned. Our work directly builds on this newly proposed approach which demonstrates substantive improvements over ACT-based methods, but remains tailored to static images and has only been validated on synthetic examples. We propose a temporally structured neural field representation for the reconstruction of auroral videos.

\subsection{Dynamic Neural Fields}

Several approaches have extended neural fields to dynamic scenes. Feeding time as an additional input to the MLP is an immediate parametrization method for continuous time varying fields \cite{gaoMonocularDynamicView2022, liNeuralSceneFlow2021, xianSpacetimeNeuralIrradiance2021}. In that case, the weights of the neural network encode the entire spatiotemporal field. For dynamic realistic scenes, an alternative approach consists instead in jointly optimizing two neural networks: the first learns a canonical representation of the field at a fixed time while the second learns a time-dependent varying deformation field \cite{parkNerfiesDeformableNeural2021,parkHyperNeRFHigherdimensionalRepresentation2021,pumarolaDNeRFNeuralRadiance2021,tretschkNonRigidNeuralRadiance2021}. A hybrid combination that jointly optimizes a temporally varying density and velocity field have also been proposed in a physics inverse problem setting \cite{fengDynamicBlackholeEmission2026}.
Instead of continuous-time inputs, per-frame latent codes have also been employed \cite{gafniDynamicNeuralRadiance2021,kimSyncNeRFGeneralizingDynamic2024}.

We adopt this last representation. A continuous time input imposes a strong smoothness prior on the reconstructed temporal evolution, while auroras can present high frequency temporal variations. A deformation field formulation is also inappropriate in our case, as auroras deform, appear or disappear between consecutive frames. We additionally model the discrete time evolution by a learnable non-linear state space process.
\section{Forward Model}

We introduce the physical forward model that we aim to invert. We first formally define the precipitating electron flux, then how it maps to 3D auroral emissions, and finally how these emissions are projected into camera observations.

\subsection{Precipitating Electron Flux}

We employ the standard assumptions used in aurora tomography \cite{tanakaFeasibilityStudyGeneralizedAurora2011} and consider a fixed volume of the atmosphere \(\volume = [x_\tmin, x_\tmax]\times[y_\tmin, y_\tmax]\times[z_\tmin, z_\tmax]\) 
where the geomagnetic field is assumed uniform. Coordinates are expressed in a field-aligned \textit{flux-tube reference frame}, where the upward \(z\)-axis is aligned with the direction of the geomagnetic field.
In this reference frame, the emission of photons at a position \((x,y,z)\in\volume\) is caused by the precipitating electrons entering the same flux tube through the top boundary at \((x,y,z_\tmax)\).

Our objective is to recover the downward \textit{differential number flux} \(\flux\) [\(\mr{m}^{-2}\mr{s}^{-1}\mr{eV}^{-1}\)]: at a horizontal position \(\postop\in\surface = [x_\tmin,x_\tmax]\times[y_\tmin,y_\tmax]\), \(\flux(x,y,E)\dif E\) is the number of electrons per unit area and unit time with energies in \([E,E+\dif E]\) that enter the volume at \((x,y,z_\tmax)\). The interaction between these electrons and the constituents of the atmosphere produces photons at specific wavelengths. Electrons with higher energies travel farther in the atmosphere, causing the emission of light at lower altitudes. We refer to \cref{fig:volume_visualization} for a visualization.

The produced 3D emission field is observed by a network of ground-based all-sky cameras at discrete times \(t_k\), \(k \in \intrange{1}{T}\). The cameras record the integrated volume emission rate (VER) at the standard auroral lines at the optical wavelengths of 427.8, 557.7, and 844.6 nm. To increase brightness and reduce noise, auroral imaging systems typically operate with an exposure time \(\Delta t = t_{k+1} - t_k\) of 1 to 2 s \cite{ogawaDevelopmentLowcostMultiwavelength2020}. We therefore do not attempt to resolve variations of the precipitating flux within an exposure, and instead assume that the flux is piecewise constant over each shutter interval and reconstruct a sequence of fields \(\flux_1, \dots, \flux_T\).

\subsection{Emission Model}

For a wavelength \(\lambda\), the instantaneous steady-state VER \(\volrate_{\mr{ss},\lambda}\) [\(\mr{m}^{-3}\mr{s}^{-1}\)] at a position \(\pos=(x,y,z)\in\volume\) relates to the electron flux at \((x,y,z_\tmax)\) via:
\begin{equation}\label{eqn:ss_volrate}
\volrate_{\mr{ss},\lambda}(\pos)
=
\int_{\mathcal{E}}
\kernel_\lambda(z,E)
\flux(x,y,E)
\dif E ,
\end{equation}
where \(\mathcal{E} = [E_\tmin,E_\tmax]\) is the reconstructed energy range, and \(\kernel_\lambda(z,E)\) is the differential emission kernel which describes the expected production of photons of wavelength \(\lambda\) at altitude \(z\) caused by precipitating electrons with energy \(E\).  We use the emission kernels derived from the GLOW model \cite{solomonGlobalModelingThermospheric2017} for our experiments. To simplify notations, we omit the dependence of the wavelengths in the following discussions. Under our piecewise constant flux assumption, the steady-state VER at a point \(\pos\) is constant during the shutter interval of frame \(k\) with value \(L_{\mr{ss},k}(\pos)\) and produced by \(\flux_k\).

\subsection{Delayed Emissions}

Some auroral emissions arise from metastable excited states with finite effective lifetimes \cite{nozawaNewFivewavelengthPhotometer2018}, thus the observed emission is a temporally delayed response to the steady-state emission rate.
For a wavelength with lifetime \(\tau\), the VER \(L\) is a first order response to the steady-state emission:
\begin{equation}
    \tau \dot{L} + L = L_\mathrm{ss}.
\end{equation}

The 427.8 nm and 844.6 nm emissions have lifetimes below 60 ns, so we assimilate their VER averaged over the exposure interval at frame \(k\) to the steady-state VER: \(L_k = L_{\mathrm{ss},k}\).
In contrast, the 557.7 nm green-line emission has a lifetime of approximately 0.7 s. Therefore, emissions at this wavelength observed during the shutter interval can depend on the precipitation occurring before the beginning of that frame.
We approximate the exposure averaged VER of the green-line emissions at frame \(k\) as:
\begin{equation}\label{eqn:decayed_emis}
    L_k = w_0 L_{\mathrm{ss},k}
    + w_1 L_{\mathrm{ss},k-1}
    + w_2 L_{\mathrm{ss},k-2}.
\end{equation}
Here, \(w_0\) weighs the contribution generated during the current exposure, \(w_1\) the residual emission from the preceding frame, and \(w_2\) the remaining older emission history. For \(\tau = 0.7\) s and an exposure time of 2 seconds, \(w_0\), \(w_1\) and \(w_2\) are respectively \(0.6701\), \(0.3110\) and \(0.0189\).

\subsection{Image Formation Model}

We approximate the photon flux intensity received by a pixel as the integrated VER along the ray aligned with its central line of sight.
Assuming isotropic emissions, the photon flux intensity received by a pixel with origin \(\ro\in\R^3\) and unit view direction \(\rd\in\R^3\) is given by:
\begin{equation}\label{eqn:pixel_int}
\pix(\ro, \rd)
=
\frac{T(\ro, \rd)}{4\pi}
\int_{s_n}^{s_f}
\volrate(\ro+s\rd)
\dif s ,
\end{equation}
where \(s_n\) and \(s_f\) are the near and far intersection distances between the pixel's ray and \(\volume\). The factor \(T(\ro, \rd)\) is the wavelength-dependent atmospheric transmittance calculated over the light path below the reconstruction volume, computed for a plane-parallel atmosphere from the zenith angle of \(\rd\). This is a common assumption used in ACT \cite{enellTomographylikeRetrievalAuroral2012} as auroral emissions mainly occur at altitudes above 100 km where the atmospheric attenuation is negligible. We exclude situations where clouds prevent the aurora from being observed, as they cause strong and spatially varying attenuation within and across views.
Finally, since the flux and emissions are assumed constant during an exposure, the averaged intensity measured during the exposure of frame \(k\) directly equals the intensity \(\pix\) computed from \(\flux_k\).

\section{Inversion Method}

We solve the reconstruction problem by modeling the precipitating flux field as a temporally conditioned neural network mapping a horizontal position at the top of the atmospheric volume to a continuous parametrization of the local energy distribution. We detail here the architecture of our neural network and its training objective.

\begin{figure}
  \centering
  \includegraphics[trim={0 0.25cm 0 0},clip,width=\linewidth]{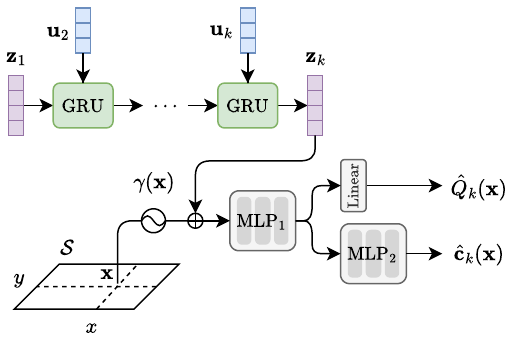}
  \caption{Visualization of our sequential neural field architecture. We iteratively construct a latent code \(\embed_k\) representing the precipitation state at frame \(k\) via a recurrent GRU unit controlled by learned innovation codes. A positionally encoded neural network decodes the frame code and input position \(\postop\in\surface\) separately into the total energy and spectral shape parameters.}
  \label{fig:nn_model}
\end{figure}

\begin{figure*}
  \centering
  \includegraphics[width=\linewidth]{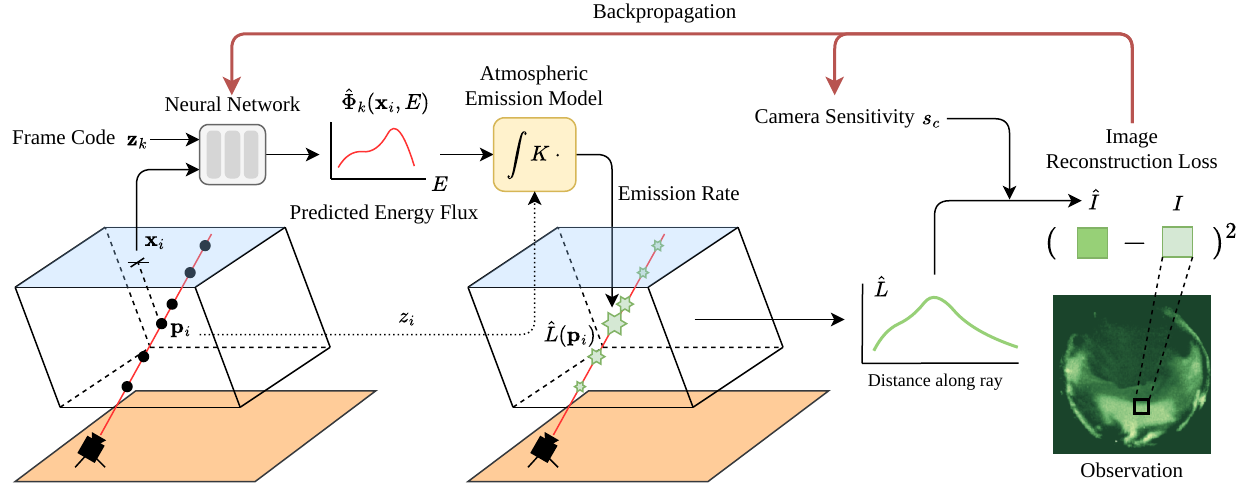}
  \caption{Overview of our optimization scheme. We synthesize observed images by sampling 3D positions within the atmospheric volume along camera rays. Each sampled position is projected to its corresponding 2D location at the top of the volume and fed into the neural precipitation field, conditioned by a frame code \(\embed_k\), which predicts the energy spectra at each location. These spectra are mapped to light emissions at the original 3D positions via a physical emission model, which are then integrated into the ray's pixel intensity. Estimated camera sensitivities scale the predicted pixel values. The entire forward process being differentiable, we optimize the neural network weights and camera sensitivities jointly through backpropagation.
  }
  \label{fig:reconstruction_pipeline}
\end{figure*}

\subsection{Flux Parametrization}

We factorize the differential energy flux \(\flux(\postop, E)\) into a total energy flux amplitude \(\eflux(\postop)\) [\(\mr{W}/\mr{m}^2\)], similarly to standard auroral precipitation parameterizations \cite{robinsonCalculatingIonosphericConductances1987}, and a normalized spectral shape \(\phi(E; \postop)\):
\begin{equation}\label{eqn:flux_parametrization}
    \flux(\postop, E) = \eflux(\postop)\phi(E; \postop)
\end{equation}
The spectral shape is represented by a fixed function basis in log-energy and log-flux space:
\begin{equation}\label{eqn:shape_parametrization}
    \phi(E; \postop) = \frac{1}{Z_k(\postop)}\exp \sum_{i=1}^\nbasis c_{i}(\postop)B_i(\log E)
\end{equation}
where the normalization factor \(Z_k(\postop)\) is computed so that \(\int_\erange \phi(E;\postop)E\dif E = 1\)
to ensure that \(\eflux(\postop)\) matches the physical total energy at \(\postop\).
We choose a B-spline basis for the \(B_i\) functions to maintain a flexible representation that does not assume a fixed spectral family (such as Maxwellian or Kappa distributions). Parameterizing log-flux rather than flux ensures positivity and allows a low-dimensional basis to remain well-conditioned as energy spectra typically exhibit variations over several orders of magnitude.

\subsection{Neural Field Representation}
\label{sec:neural_field}

We model the reconstructed parametrization \(\hat\eflux_k\) and \(\hat{\mathbf{c}}_k = [c_{k,1}, \dots, c_{k,\nbasis}]^\top\) at frame \(k\)\ and position \(\postop\) as the outputs of a neural network conditioned on a latent code \(\embed_k \in \R^d\):
\begin{equation}
\begin{aligned}
    &\mathbf{h}_k(\postop) = \mathrm{MLP}_1([\gamma(\postop), \embed_k])\\
    &\hat\eflux_k(\postop) = \exp\left(\mathbf{w}^\top \mathbf{h}_k(\postop) + b\right)\\
    &\hat{\mathbf{c}}_k(\postop) = \mathrm{MLP}_2(\mathbf{h}_k(\postop))
\end{aligned}
\end{equation}
where \(\gamma\) denotes a positional encoding that maps an input 2D position into higher dimensional Fourier basis \cite{tancikFourierFeaturesLet} of increasing frequencies:
\begin{equation}
\begin{aligned}
    \gamma(\postop) = [
        &\postop, \sin(\pi\postop), \cos(\pi\postop),\\
        &\dots,
        \sin(2^{l-1}\pi\postop), \cos(2^{l-1}\pi\postop)
    ]^\top.
\end{aligned}
\end{equation}
A higher degree \(l\) allows the MLP to represent higher frequency spatial features.
We choose a latent code representation over a continuous-time MLP that would directly take \((\postop, t)\) as input to better match the sequential structure of the recorded images, which can contain rapidly varying spatial structures that appear or disappear from one frame to the next.

Learning \(T\) independent codes \(\embed_k\) would however poorly exploit the redundancy of structures that vary slowly.
We therefore model the sequence of temporal codes as the result of a recurrent non-linear state-space process. We store \(T - 1\) learnable \textit{innovation} codes \(\innov_2, \dots, \innov_T \in\R^{d_i}\), with \(d_i < d\), and use a single-layer gated recurrent unit (GRU) \cite{choLearningPhraseRepresentations2014} to iteratively construct \(\embed_{k+1}\) from \(\embed_k\) and \(\innov_{k+1}\). The code \(\embed_1\) of the first frame remains a free learnable parameter. A diagram of our proposed architecture is shown in \cref{fig:nn_model}.
Our experiments show that this recurrent temporal modeling yields better reconstruction than simply regularizing \(\|\embed_{k+1} - \embed_{k}\|\).

\subsection{Optimization Objective}

We optimize the network parameters and latent codes to minimize the loss:
\begin{equation}
\loss =  \sum_\lambda \loss_{\mr{img},\lambda} + \mu_\mr{log}\loss_\mr{log} + \mu_\mr{lin}\loss_\mr{lin}.
\end{equation}
where \(\loss_{\mr{img},\lambda}\) measures the discrepancy between the ground truth observations and the synthesized views from the modeled fields \(\hat\flux_1,\dots,\hat\flux_T\), and \(\loss_\mr{log}\) and \(\loss_\mr{lin}\) are regularization terms imposing a smoothness prior on the energy spectra. The following paragraphs describe each loss term.

We measure the image reconstruction error at frame \(k\) by comparing the modeled intensity \(\hat\pix_{p,k}\) of each pixel \(p\) against the observed value \(\pix_{p,k}\) by the cameras.
The observed pixel value is subject to standard read and shot noise whose effect is emphasized by the high dynamic range of the photon flux that reaches the camera sensor.
We therefore define the reconstruction loss \(\loss_{\mr{img},\lambda}\) as a pixel variance-weighted mean squared error across all frames:
\begin{equation}
\loss_\mr{img,\lambda} = \frac{1}{|\mathcal{R}_\lambda|}\sum_{ \substack{p\in\mathcal{R}_\lambda\\k\in\intrange{1}{T}}}
\frac{(\hat\pix_{p,k} - \pix_{p,k})^2}{\alpha \mr{sg}(\hat\pix_{p,k}) + \beta}
\end{equation}
where \(\mathcal{R}_\lambda\) is the set of pixels whose view direction intersects \(\volume\), and \(\mr{sg}\) denotes the stop gradient function. The denominator matches standard noise modeling where the variance is proportional to the mean pixel value by a factor \(\alpha\) value and offset by a noise floor \(\beta\).

We impose the same smoothness prior on the energy spectra by penalizing the spectral roughness, which quantifies the second order variations of \(\hat\flux(\postop, E)\). \(\loss_\mr{log}\) regularizes the roughness in log-energy log-flux space, as proposed by \cite{misyatsFastScalableReconstruction2026}, while \(\loss_\mr{lin}\) regularizes the spectra in linear space as used in ACT-based reconstruction \cite{fukizawaThreedimensionalIonosphericConductivity2023}.
Log-domain regularization is scale-invariant and prevents the formation of high frequency components over the several orders of magnitude of the energy range. The linear-domain complementarily suppresses large dynamic range spikes that the log term is blind to.

\subsection{Ray Integration with Radiometric Calibration}

Our optimization requires to repeatedly render the projected emissions \(\hat\pix_{p,k} = \hat\pix_{k}(\ro_p, \rd_p)\) for each pixel \(p\) in a differentiable manner.
To do so, we draw \(N\) points along the segment of ray intersecting \(\volume\) using stratified sampling as in NeRFs \cite{mildenhallNeRFRepresentingScenes2021}. For each sampled point \(\pos_i = (x_i, y_i, z_i)\) along the ray, we evaluate the spectrum \(\hat\flux_k\) at the projected 2D position \((x_i, y_i)\). We then numerically integrate equation \eqref{eqn:ss_volrate} using a pre-computed piecewise-constant emission kernel over \(50\) altitude and \(50\) energy bins, yielding \(\hat\volrate_{\mr{ss},k}(\pos_i)\). For the 557.7 nm wavelength, this procedure is repeated two more times with \(\hat\flux_{k-1}\) and \(\hat\flux_{k-2}\) over the same \(N\) points. The pixel intensity \(\hat\pix_{p,k}\) is finally obtained by numerically integrating \(\volrate_k(\pos_1)\), ..., \(\volrate_k(\pos_N)\) over the ray interval.

In practical reconstruction of real aurora recordings, cameras are often partially calibrated and their relative sensitivities are unknown. ACT estimates the sensitivity of each camera through cross-validation on the reconstructed images \cite{fukizawaThreedimensionalIonosphericConductivity2023}, which requires repeating the entire reconstruction process multiple times. We instead learn the sensitivities of each camera jointly with the reconstruction, assuming at least one camera per wavelength is calibrated. We store for each non-calibrated camera \(c\) a learnable sensitivity parameter \(\alpha_c\) which maps to the camera sensitivity factor \(s_c = \exp \alpha_c\). This factor is fixed to 1 for calibrated cameras. During rendering, the calculated intensities \(\hat\pix_{p,k}\) are scaled by their respective camera sensitivities before entering the loss. \cref{fig:reconstruction_pipeline} shows a visualization of the overall rendering and optimization pipeline.

\section{Experiments and Results}

We first evaluate the proposed method on two synthetic auroral sequences for which the precipitating electron flux is known. We report reconstruction accuracy for the total energy flux, mean energy and spectral shape. We then evaluate the method on a real auroral recording through comparisons with ACT. Further discussion and details are available in the supplementary material.

\subsection{Simulated Auroras}

\paragraph{Experimental setup.}
The reconstruction domain is \(\volume=[-75,75]\times[-100,100]\times[80,180]\) km in south, east and altitude coordinates centered near Tromsø. The modeled energy interval is \(\erange=[0.3,50]\) keV. Synthetic observations are generated at Tromsø, Skibotn and Kilpisjärvi.
Each camera channel produces \(640\times480\)-pixel images with a 2 s exposure at 427.8, 557.7 and 844.6 nm, except the Tromsø 844.6 nm channel, which uses 256 × 256 pixels as in the existing setup. A sensitivity sampled uniformly from \([0.5,1.5]\) is assigned to each site and wavelength pair, with Tromsø fixed to 1. We add signal-dependent Gaussian noise with variance \(4I+40^2\). Exposure integration and the finite lifetime of the 557.7 nm emission are included. A pre-processing step removes uniform background emissions.
Further details are in the supplementary material.

\paragraph{Evaluation metrics.}
We use the total energy flux \(\eflux\) [mW/m\(^{2}\)] and mean energy \(\emean\) [keV] as metrics for the reconstruction accuracy, respectively defined, up to a unit scaling factor, as:
\begin{equation}
  \eflux(\postop) = \int_\erange \flux(\postop, E)E\dif E,
  \ 
  \emean(\postop) = \frac{
    \int_\erange \flux(\postop, E)E\dif E
  }{
    \int_\erange \flux(\postop, E)\dif E.
  }
\end{equation}    
The total energy flux -- modeled directly by our parametrization \eqref{eqn:flux_parametrization} -- correlates with the brightness of the emissions, while the mean energy locates the center of mass of the energy distribution.

\paragraph{Synthetic datasets.}

\begin{figure}
    \centering
    \includegraphics[width=1\linewidth]{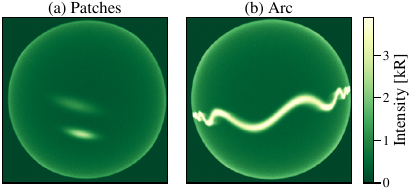}
    \caption{Sample observations of the two synthetic auroras viewed from Tromsø at the 557.7 nm wavelength.}
    \label{fig:synthetic_observations}
\end{figure}

We simulate simplified models of the two typical types of auroras: diffuse and discrete.
The diffuse aurora consists of two Gaussian-shaped auroral patches with a Maxwellian energy spectra. The first patch remains fixed throughout the observation time while the second repeatedly drifts and fades in intensity over a 10 s cycle. The simulated aurora is observed over 1 minute yielding 30 images. A sample synthetic observation is shown in \cref{fig:synthetic_observations} (a).
The discrete aurora consists in a bright sinusoidal arc traveling westward at 1 km/s with a precipitating flux consisting of two power laws and a Gaussian peak whose center oscillates between 3.5 and 6.5 keV over a 100 s cycle. This arc is recorded over 5 minutes producing 150 images per camera. \cref{fig:synthetic_observations} (b) shows an observation of this arc.

\paragraph{Results.}
We run our reconstruction on both synthetic datasets over 10K iterations with 1024 rays per batch, 64 samples per rays and regularization strength \(\mu_\mr{log} = 100\) and \(\mu_\mr{lin} = 0\) which we found empirically to produce the best results. Frame embeddings and innovation codes have dimension \(d = 16\) and \(d_i = 4\) respectively. The first and second MLPs contain respectively 3 and 2 hidden layers separated with ReLU activations. The reconstruction time was 3 to 5 minutes on an NVIDIA RTX 4090 GPU.

\begin{figure}
    \centering
    \includegraphics[width=\linewidth]{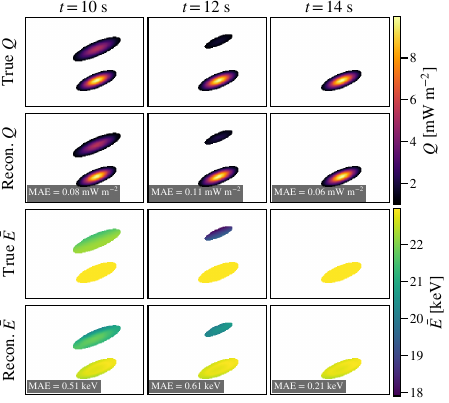}
    \caption{Comparison between the reconstructed and ground truth spatially varying total energy flux \(\eflux\) and mean energy \(\emean\) of the synthetic patches dataset at three arbitrarily selected frames over the reconstructed 2D spatial domain at the top of the atmospheric volume. Only regions where \(Q > 1\) mW/m\(^2\) are shown.}
    \label{fig:patches_reconstruction}
\end{figure}

\begin{figure}[t]
    \centering
    \begin{subfigure}{\columnwidth}
        \captionsetup{position=top}
        \caption{Energy spectra at the center of each patch at \(t\) = 12 s.}
        \centering
        \includegraphics[width=\linewidth]{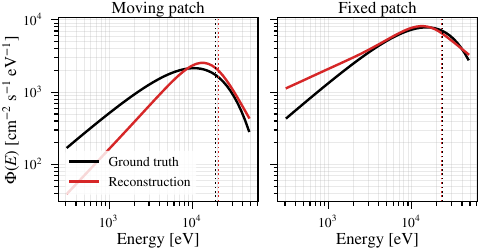}
        \label{fig:patches_spectra}
    \end{subfigure}
    \vspace{0.5em}
    \begin{subfigure}{\columnwidth}
        \captionsetup{position=top}
        \caption{Energy spectra in the middle of the simulated discrete arc.}
        \centering
        \includegraphics[width=\linewidth]{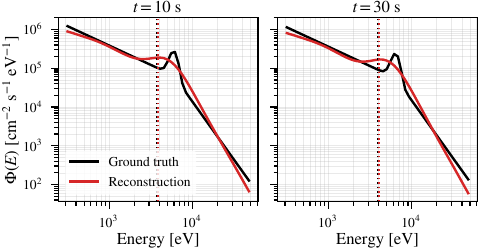}
        \label{fig:arc_spectra}
    \end{subfigure}
    \caption{Comparison between the reconstructed and ground truth energy spectra at select locations of our two synthetic datasets. The dashed vertical lines locate the reconstructed (red) and true (black) mean energy \(\emean\).}
    \label{fig:synthetic_spectra}
\end{figure}

\cref{fig:patches_reconstruction} shows that the reconstruction
accurately recovers the position, spatial extent and intensity of the compact patches. It also captures the
appearance and fading of the pulsating patch between consecutive
exposures. The reconstructed spectra in
\cref{fig:synthetic_spectra} recover the dominant energy distribution
and its temporal evolution. The main discrepancy occurs for
the narrow spectral peak of the discrete arc, which is reconstructed
with a smoother profile. Nevertheless, its
time-varying mean energy is recovered accurately.
Finally, \cref{tab:sensitivities} demonstrates that our joint reconstruction and calibration scheme allows to effectively estimate relative camera sensitivities. 

\begin{table}[htbp]
\centering
\begin{tabular}{llccc}
\toprule
 & & True & \multicolumn{2}{c}{Est. sensitivity} \\
\cmidrule(lr){4-5}
Site & Wavelength & sensitivity & Patches & Arc \\
\midrule
Skibotn & 427.8 & 1.27 & 1.25 & 1.26 \\
Skibotn & 557.7 & 0.81 & 0.81 & 0.81 \\
Skibotn & 844.6 & 1.40 & 1.34 & 1.37 \\
Kilpisj\"{a}rvi & 427.8 & 0.59 & 0.58 & 0.57 \\
Kilpisj\"{a}rvi & 557.7 & 1.13 & 1.14 & 1.13 \\
Kilpisj\"{a}rvi & 844.6 & 0.96 & 0.93 & 0.91 \\
\bottomrule
\end{tabular}
\caption{Comparison between ground truth camera sensitivities and estimated sensitivities during reconstruction on the synthetic datasets.}
\label{tab:sensitivities}
\end{table}

\subsection{Real Observations}

\begin{figure}[t]
    \centering
    \includegraphics[width=\linewidth]{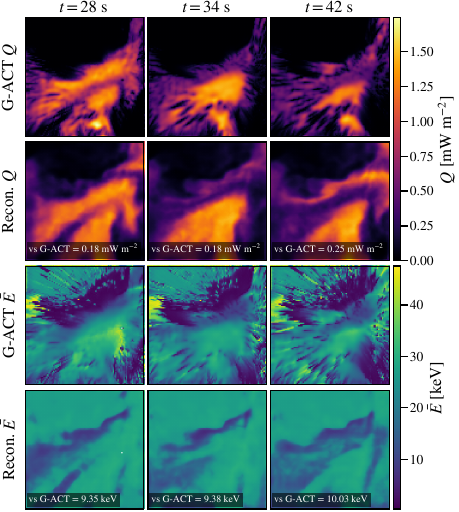}
    \caption{Reconstructed total energy flux and mean energy by G-ACT and our method at selected frames. Numerical comparisons are the mean absolute difference from G-ACT. Large errors in mean energy can be attributed to low energy flux regions where it is not well defined.}
    \label{fig:real_reconstruction}
\end{figure}

We apply our method on a real aurora observed on 26 March 2023 and compare against the reconstruction obtained by G-ACT on a sequence of 10 frames.
We increase the dimension of the frame and innovation codes to 32 and 16 dimensions, linear regularization to \(\mu_\mr{lin} = 10^5\), and 20K iterations. All images were pre-processed as described in \cite{fukizawaThreedimensionalIonosphericConductivity2023}. In particular, camera calibrations are already compensated.
\cref{fig:real_reconstruction} shows the reconstructed mean and total energy flux of both G-ACT and our method, showing that our neural field yields a smoother spatial distribution. The large linear regularization was required to produce consistent reconstruction with G-ACT, but yielded particularly smooth energy spectra, which may differ significantly from the real energy spectra and discuss this further in the supplementary material. 
\cref{fig:camera_renders} compares re-rendered views from G-ACT and our method.

\begin{figure}[t]
    \centering
    \includegraphics[width=\linewidth]{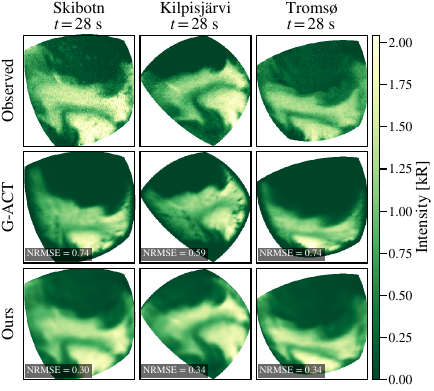}
    \caption{Comparison between camera observations (top), images rendered with the reconstructed flux from G-ACT (middle), and images obtained from our reconstruction (bottom). NRMSE is calculated as RMSE divided by the standard deviation of the observed pixels. All three views were used during optimization. Our reconstruction technique reproduces each frame closer to the observations than G-ACT.}
    \label{fig:camera_renders}
\end{figure}

\subsection{Temporal Ablation}

As mentioned in \cref{sec:neural_field}, we test whether our learnable dynamic state-space process via a GRU provides improved reconstruction over independent per-frame latent code or directly feeding the continuous frame time as an input to our neural field.
More precisely, we compare three architectures against our GRU-based method: independently learned per-frame codes, without and with an additional regularization term penalizing the difference between each code, and a continuous time model.
For the continuous time model, the input time is the normalized frame index passed through a separate Fourier basis encoding than the input position. 
We use 8 and 4 time harmonics for the patches and arc, respectively.
\cref{tab:patches_temp_ablation} shows that the GRU outperforms other temporal conditioning on the total energy flux for both datasets and on arcs for the mean energy. Independent codes are slightly better for the patches.

\begin{table}[]
\centering
\begin{tabular}{lcccc}
\toprule
 & \multicolumn{2}{c}{$Q$ MAE $\downarrow$} & \multicolumn{2}{c}{$\bar{E}$ MAE $\downarrow$} \\
\cmidrule(lr){2-3} \cmidrule(lr){4-5}
Model & Patches & Arc & Patches & Arc \\
\midrule
Independent & 0.087 & 1.459 & \textbf{0.411} & 1.265 \\
Difference reg. & 0.096 & 1.423 & 0.521 & 1.842 \\
Continuous & 0.081 & 0.274 & 0.466 & 0.127 \\
GRU & \textbf{0.073} & \textbf{0.249} & 0.417 & \textbf{0.120} \\
\bottomrule
\end{tabular}
\caption{Mean absolute error of the reconstructed total energy flux and mean energy averaged across all frames of both simulated datasets, and reconstructed with different temporally conditioned neural fields.}
\label{tab:patches_temp_ablation}
\end{table}
\section{Conclusion}

We proposed a temporally conditioned neural field representation for the reconstruction of the precipitating electron flux causing auroras. Our technique recovers the spatially and temporally varying energy spectra of electrons at the top of the atmosphere, showing accurate reconstruction on synthetic data, and consistent results with existing methods on real events. Tuning remains a particular challenge of our method. Further experimentation is needed to assess correct optimization parameters and potential biases in the reconstructed flux. Higher time resolution recordings could also be employed to better resolve auroral dynamics.
Additionally, our neural representation can also be integrated with other sensor measurements such as scattering radars, which would further constrain and validate the reconstruction.

Most importantly, our method enables significantly faster reconstruction than conventional tomography, making it possible to investigate the spatiotemporal evolution of auroral electron precipitations using the large volume of observations accumulated over past decades.
The statistical analysis of multiple auroral events could help advance our understanding of the physical mechanisms governing electron precipitation from the magnetosphere into the ionosphere.
Combined with additional sensors, it would offer the opportunity to study the three-dimensional structure of ionospheric current systems associated with auroral activity.

\section*{Acknowledgments}

This research was supported by ``Strategic Research Projects'' grant from ROIS (Research Organization of Information and Systems).

{
    \small
    \bibliographystyle{ieeenat_fullname}
    \bibliography{main}
}

\clearpage
\appendix

\maketitlesupplementary
\appendix
This supplementary material provides additional details on the forward
model, network architecture, training procedure on the synthetic and real datasets, as well as additional qualitative results.
Section A describes the derivations of the weights for the 3-frame decay model used for the 557.7 nm wavelength emissions. Section B provides additional details on the synthetic datasets and their reconstruction. Section C describes the background removal procedure. Finally, Section D provides a discussion on the reconstruction of the real dataset.

\section{Weights of the 557.7 nm emissions}

We derive the coefficient \(w_0\), \(w_1\) and \(w_2\) used to approximate the finite lifetime of the 557.7 nm wavelength.
The continuous-time volume emission rate (VER) \(L\) relates to the steady-state VER \(L_\mr{ss}\) via
\begin{equation}\label{eqn:L_ode}
    \tau \dot L + L = L_\mr{ss}
\end{equation}
which has the closed form solution:
\begin{equation}\label{eqn:L_integral}
    L(t) = \frac{1}{\tau}\int_{-\infty}^t L_\mr{ss}(t')e^{-(t-t')/\tau}\dif t'.
\end{equation}
The cameras capture frames at discrete times \(t_k = k\Delta t\) with exposure time \(\Delta t\). We assume the precipitating flux to be piecewise constant over the frame shutter interval \([t_k, t_{k+1}]\).
This implies that the steady-state VER is also constant over this same interval, \ie \(L_\mr{ss}(t') = L_{\mr{ss},k}\) for \(t' \in [t_k, t_{k+1}]\).
The averaged VER \(L_k\) of \(L\) over \([t_k, t_{k+1}]\) can then be expressed as:
\begin{equation}\label{eqn:ver_avg_int}
\begin{aligned}
    L_k &= \frac{1}{\Delta t}\int_{t_k}^{t_{k+1}} L(t)\dif t
    =
    \sum_{j=-\infty}^{k} a_{k,j} L_{\mr{ss},j}
\end{aligned}
\end{equation}
where
\begin{equation}\label{eqn:akj}
\begin{aligned}
    a_{k,j} &= \frac{1}{\tau\Delta t} \int_{t=t_k}^{t_{k+1}} \int_{t'=t_j}^{\min(t_{j+1},t)} e^{-(t-t')/\tau}\dif t\dif t'.
\end{aligned}
\end{equation}

For an exposure time of \(\Delta t = 2\) s and an effective lifetime of \(\tau = 0.7\) s, we have \(\exp(-(t-t')/\tau) < 3.3\times 10^{-3}\) when \(t - t' > 4\) s. We therefore neglect the contributions of emissions during frames \(k-3\) and before and approximate the expression of \(L_k\) in \eqref{eqn:ver_avg_int} as:
\begin{equation}
    L_k = w_0 L_{\mr{ss},k} + w_1 L_{\mr{ss},k-1} + L_{\mr{ss},k-2}\underbrace{\sum_{j=-\infty}^{k-2} a_{k,j}}_{w_2}.
\end{equation}
Computing \(w_0\), \(w_1\) and \(w_2\) using \eqref{eqn:akj} yields:
\begin{equation}
\begin{aligned}
    w_0 &= 1 - \frac{\tau}{\Delta t}\left(1 - e^{-\Delta t/\tau}\right)\\
    w_1 &= \frac{\tau}{\Delta t}\left(1 - e^{-\Delta t/\tau}\right)^2\\
    w_2 &= \frac{\tau}{\Delta t}e^{-\Delta t/\tau}\left(1 - e^{-\Delta t/\tau}\right).\\
\end{aligned}
\end{equation}
Evaluating them numerically gives
\(w_0 \approx 0.6701\), \(w_1 \approx 0.3110\) and \(w_2 \approx 0.0189\).
\section{Synthetic Experiments}

We provide here a more detailed description of our synthetic dataset and the training procedure on these datasets.

\subsection{Camera Configuration}

The three imaging stations, Tromsø and Skibotn, Norway, and Kilpisjärvi, Finland, and the two wavelengths 427.8 nm and 557.7 nm mirror the real multi-station all-sky camera network used for the 26 March 2023 event and by the G-ACT reference method \cite{fukizawaReconstructionPrecipitatingElectrons2022, fukizawaThreedimensionalIonosphericConductivity2023}. The two emission lines are commonly employed because of their brightness and because both respond to different parts of the precipitating electron energy spectrum and thus have different altitude-dependent emission profiles. Combining them constrains the reconstructed energy flux and mean energy better than either line alone. A third line (844.6 nm) is also simulated in the synthetic datasets as some locations are equipped with this wavelength.

The camera used in our experiments is the Watec WAT-910HX/RC, equipped with
a Sony ICX428ALL CCD sensor and a FUJINON fisheye lens (YV2.2×1.4A-SA2).
The system has a field of view of approximately 185° (about 0.4° per pixel on
average), an aperture setting of approximately F2.2, and an 8-bit output resolution.
The camera supports exposure times ranging from \(10^{-5}\) s to 4.27 s, while the exposure times typically used for auroral observations are between 0.25 and 4.27 s \cite{ogawaDevelopmentLowcostMultiwavelength2020}.
The camera system was radiometrically calibrated using a National Institute of Standards and Technology-traceable integrating sphere \cite{ogawaOpticalCalibrationSystem2020}, including gamma and flat-field correction. However,
the sensor's full-well capacity and dark current in electrons are not available in the published specifications.

\subsection{Simulated Auroras}

\begin{figure*}[t]
    \centering

    \begin{subfigure}[t]{0.49\textwidth}
        \centering
        \includegraphics[width=1\linewidth]{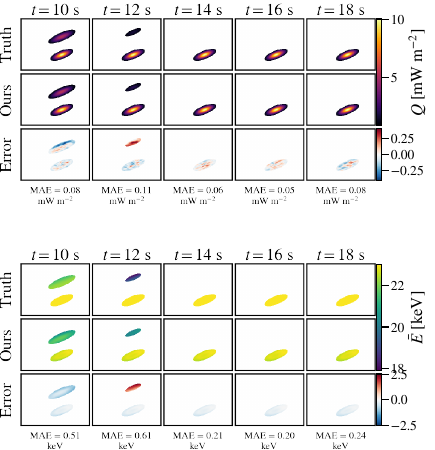}
        \caption{Patches results.}
        \label{fig:patches_reconstruction_supp}
    \end{subfigure}
    \hfill
    \begin{subfigure}[t]{0.49\textwidth}
        \centering
        \includegraphics[width=1\linewidth]{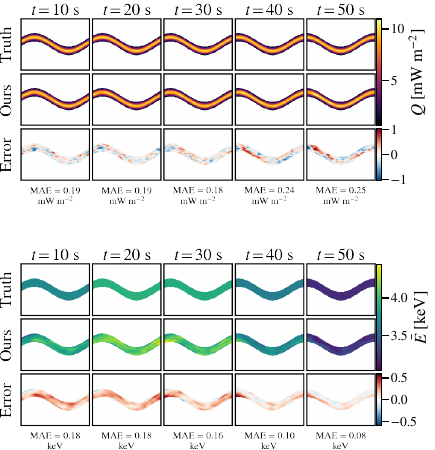}
        \caption{Arc results.}
        \label{fig:arc_reconstruction_supp}
    \end{subfigure}

    \caption{Reconstructed total energy flux and mean energy for the synthetic patches and arc dataset. Only regions where the ground truth total energy flux is above 1 mW/m\(^2\) are shown.}
    \label{fig:reconstruction_results}
\end{figure*} 

\paragraph{Auroral patches.}
This dataset simulates two auroral patches over a 30-frame sequence modeling a pulsating aurora. Both patches follow a Maxwellian spectrum. A stationary patch maintains a constant characteristic energy of 30 keV throughout the entire sequence.
A second patch propagates northward while pulsating in brightness on a 6 s on / 4 s off duty cycle, mimicking the brightness variations of real pulsating auroral patches (the same phenomenon as the real observations used for reconstruction). Its energy falls from 30 to 10 keV over a 6 s active phase.
Both patches share a peak total energy flux of 10 mW/m\(^2\) and a Gaussian spatial profile.

\paragraph{Auroral arc.}
This second dataset consists of a single auroral arc spanning the full east–west extent of the \(200\times150\) km domain, over a 150-frame, 300-second sequence at 2 s cadence. The arc follows a sinusoidal centerline (wavelength 200 km, amplitude 20 km) that drifts westward at 1 km/s, with a Gaussian brightness profile of width 8 km peaking at a total energy flux of 10 mW/m\(^2\) at its center. Its electron energy spectrum combines a Gaussian peak with a broken power-law background (different low- and high-energy exponents on either side of the peak). The peak energy itself oscillates sinusoidally between 3.5 and 6.5 keV with a 100 s period.

\subsection{Training}

\paragraph{Neural field.} For both datasets, the neural field consists of two MLPs with 3 and 2 hidden layers respectively, with dimensions of 128 and separated by ReLU activations. The intermediate hidden embedding shared by the total energy linear decoder and spectral shape decoder has a dimension of 128. We use 32 B-splines for the spectrum's representation with uniformly spaced knots in log-domain of the energy range, which spans from 0.3 keV to 50 keV. We set 3 degrees of Fourier encoding for the positional input.

\paragraph{Optimization.}
Both datasets are trained for 10,000 iterations using the Adam optimizer, with the learning rate decayed exponentially from $5\times10^{-4}$ to $5\times10^{-5}$ over the course of training. In each iteration we sample a batch of 1,024 rays per wavelength across all nine camera views (three locations at three wavelengths), each integrated with 64 stratified samples along the ray. Rendered intensities are compared against the background-subtracted observations under a variance-weighted MSE loss, whose variance is modeled as an affine function of the predicted intensity: $\alpha \hat{I} + \beta$, with $\alpha=4$ and $\beta=40^2$ R$^2$. Spectral smoothness is enforced by a second-difference penalty on the log-normalized spectral shape, evaluated at 1,024 randomly sampled spatial points per iteration with weight \(\mu_\mr{log} =  100\). The additional linear-domain smoothness term is disabled for the synthetic datasets (\(\mu_{\mr{lin}} = 0\)), which we found empirically to over-smooth the reconstructed spectra. Per-camera sensitivities are optimized jointly with the neural field parameters, using the same optimizer and learning rate schedule, with one Tromsø camera per wavelength fixed to a sensitivity of 1 to prevent ambiguity. \cref{fig:reconstruction_results} displays the reconstruction of the total energy flux and mean energy over the 2D horizontal domain for both datasets.

\paragraph{Performance.}

We run both training procedures on a computer equipped with an NVIDIA RTX 4090, performing 30 to 50 training iterations per second. For the patch dataset, where training takes about three and a half minutes.
For the arc dataset, not all 150 frames of the sequence can hold in memory during training. We therefore employ a pooling techniques, where we load 50 random frames (at all wavelengths) and samples rays within this pool for 500 iterations before loading new frames. This slows down the training time on the arc dataset to five and a half minutes. This is substantially faster than ACT and G-ACT-based techniques which report in the order of an hour for the reconstruction of a single frame \cite{misyatsFastScalableReconstruction2026}.
 
\subsection{Temporal Ablation}

\begin{figure}
    \centering
    \includegraphics[width=1\linewidth]{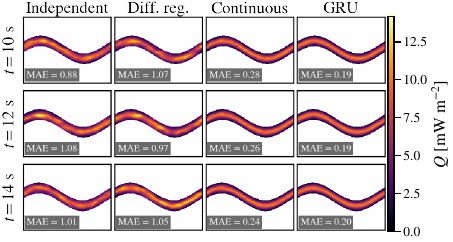}
    \caption{Reconstruction accuracy of the total energy flux on the arc dataset with different temporally-encoded neural fields.}
    \label{fig:arc_temporal_ablation}
\end{figure}

\cref{fig:arc_temporal_ablation} demonstrates the effect of the neural field's temporal representation on reconstruction quality on the arc dataset. We maintain the same architecture, loss, and training procedure as described above and vary only how each model is temporally conditioned. The model labeled \textit{independent} assigns each frame its own free embedding vector with no coupling across frames. \textit{Difference regularized} uses the same per-frame embeddings but adds an explicit penalty on the squared difference between consecutive frames' embeddings to encourage temporal smoothness. \textit{Continuous} instead conditions on a Fourier encoding of normalized time. \textit{GRU} generates the frame embeddings from a recurrent network driven by a per-frame innovation input. We evaluated the reconstruction quality of the mean energy and total energy flux at three frames of the moving arc sequence. Learnable per-frame embeddings, independent and regularized, reconstruct the spatial total energy flux and mean energy worse than the two models with a structured temporal representation. Our GRU-based neural field achieves the lowest error at all three frames, marginally ahead of the continuous-time field. Note however that the continuous-time model required manual tuning of the encoding degree (number of Fourier basis) to which it demonstrated high sensitivity. We tested encoding degrees \(l \in \{1, 2, 4, 8, 16\}\) and found that 4 produced the smallest MAE on the total energy flux.

\subsection{Comparison to G-ACT}

\begin{figure}
    \centering
    \includegraphics[width=1\linewidth]{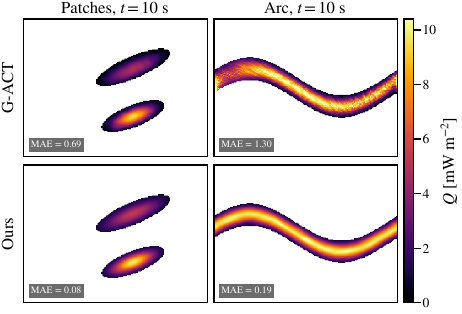}
    \caption{Total energy flux \(\eflux\) at \(t = 10\) s from G-ACT (top) and our method (bottom), on the patches and arc datasets. Only pixels with true $\eflux \geq 1 \ \mr{mW}/\mr{m}^2$ are shown. G-ACT values above the color scale are shown with the top color.}
    \label{fig:gact_comparison}
\end{figure}

We run the same G-ACT version as in \cite{fukizawaThreedimensionalIonosphericConductivity2023} on the synthetic datasets on selected frames using a \(151\times 151\) spatial grid resolution. The linear spatial and energy curvature regularization strength is set to \(0.1\) for the patches and \(10^{-5}\) for the arc. The reconstruction employs SIRT initialization followed by Gauss-Newton minimization, which was stopped when the relative residual change fell below 1\%. Each frame takes about 2 hours to reconstruct.
Note that G-ACT's forward model is steady-state: it does not include the 557.7 nm lifetime, exposure integration, or atmospheric attenuation. Images are scaled by their true sensitivities and background is removed before reconstruction. \cref{fig:gact_comparison} compares the reconstructed total energy flux obtained with G-ACT and our method at selected frames. G-ACT's mean energy MAE is 6.1 keV (patches) and 1.4 keV (arc), against 0.5 and 0.2 keV for our method.

\section{Background Removal}

\begin{figure}
    \centering
    \includegraphics[width=1\linewidth]{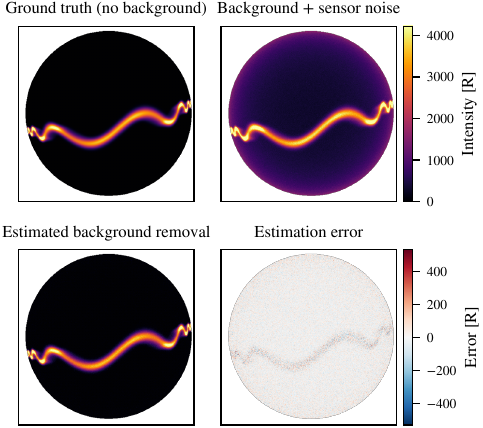}
    \caption{Result of the background removal procedure on one image of the arc dataset.}
    \label{fig:background_removal}
\end{figure}

\begin{figure*}
    \centering
    \includegraphics[width=1\linewidth]{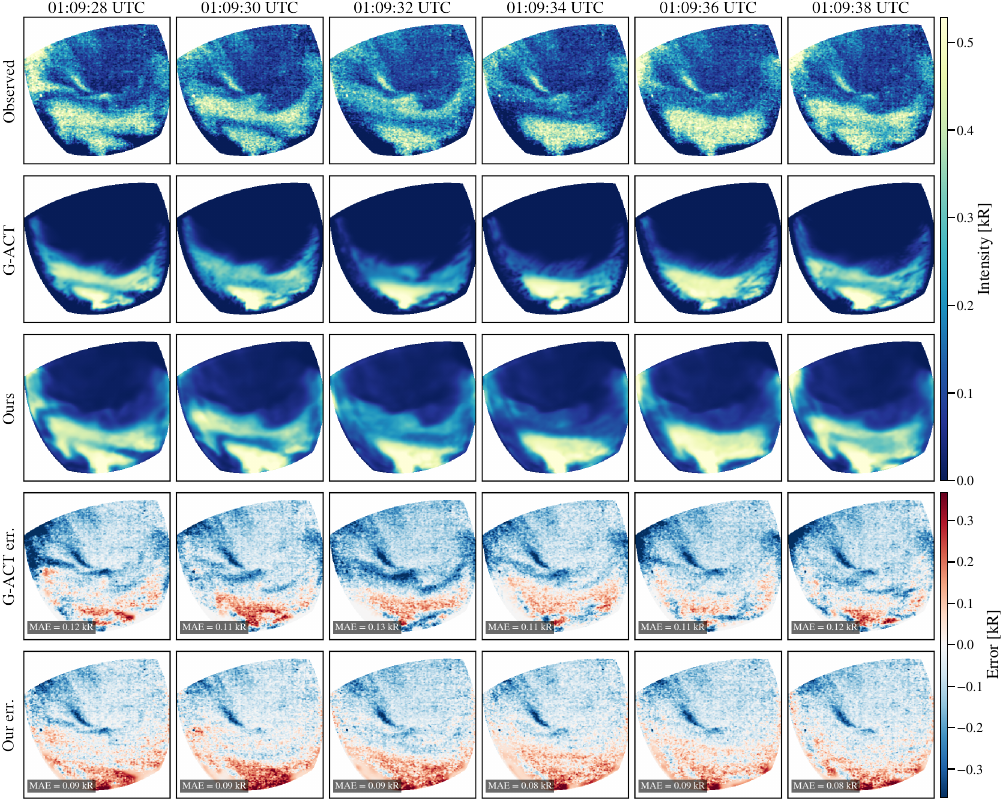}
    \caption{Re-renders of the view from Tromsø using the reconstructed flux from G-ACT and our neural field at the wavelength of 427.8 nm.}
    \label{fig:rerender_4278}
\end{figure*}

\begin{figure*}
    \centering
    \includegraphics[width=1\linewidth]{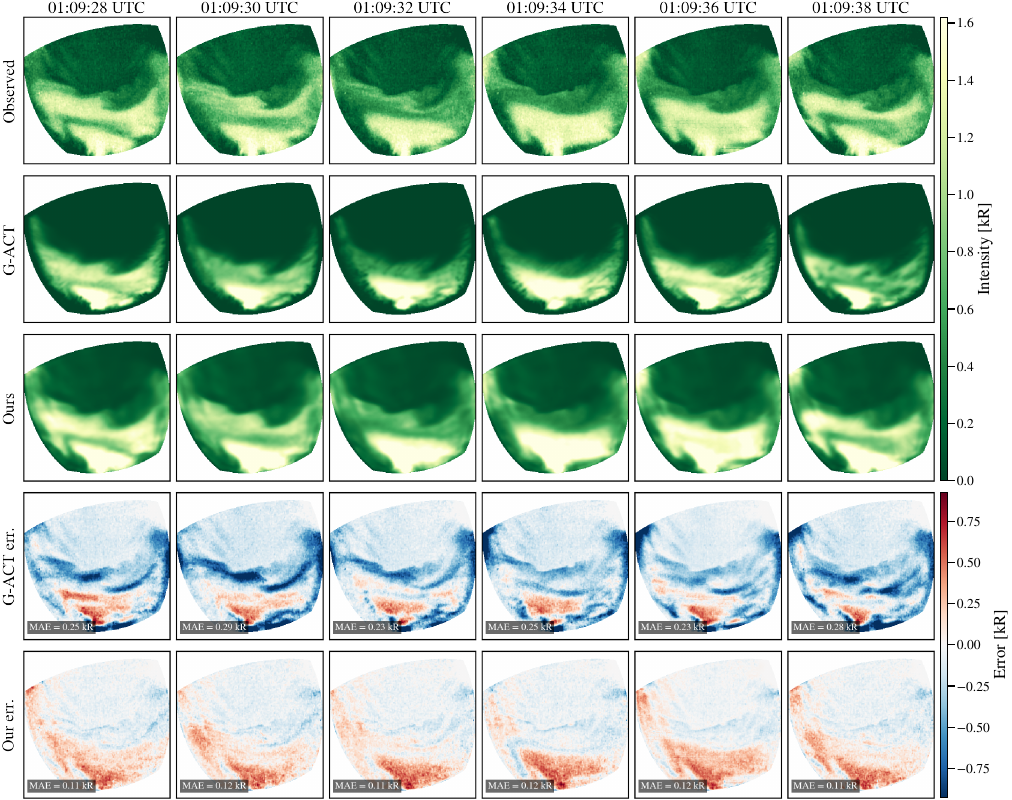}
    \caption{Re-renders of the view from Tromsø using the reconstructed flux from G-ACT and our neural field at the wavelength of 557.7 nm.}
    \label{fig:rerender_5577}
\end{figure*}

Auroral emissions are often accompanied by a spatially uniform emission
component associated with low-energy electron precipitation. For our synthetic datasets, we introduce an artificial background glow with a spatially uniform offset in the energy spectra for the arc dataset, and an additional low-energy Maxwellian tail for the patch dataset. This background
component is removed from each frame before running the reconstruction.
For real observations, the optical filter has a full width at half maximum of 10 nm, so a few stars or planets appear in the images, and the stars that do appear are removed using a median filter. Airglow is fainter than auroral
emissions and is not detectable at the exposure times of 1 to 2 s used here.
Moonlight is not a concern for these observations because the data were acquired near the new moon.

For the uniform background removal, we define \(I_\mr{bg}(\ro, \rd)\) as the intensity received by a pixel with origin \(\ro\) and direction \(\rd\) caused by a uniform emission rate of 1 cm\(^{-3}\)s\(^{-1}\) within the slice of the atmosphere bounding by our minimum and maximum reconstruction altitude.
The observed pixel $\tilde{I}(\mathbf{o}, \mathbf{d})$ (ignoring noise) is then modeled as:
\begin{equation}
\tilde{I}_\mr{bg}(\mathbf{o}, \mathbf{d}) = a + cI_\mr{bg}(\mathbf{o}, \mathbf{d}),
\end{equation}
where $c$ scales for the background's emission-rate magnitude and $a$ is an additive offset not captured by \(I_\mr{bg}\).
To find $a$ and $c$, we minimize the squared difference between the observed image $\tilde{I}$ and this model, $\sum (\tilde{I} - a - cI_\mr{bg})^2$, but only over the pixels we currently believe contain background and no aurora. We initialize this set to all pixels and then use the median absolute deviation (MAD) to refine it iteratively. More precisely, after a fit of \(a\) and \(c\), we compute the error $r = \tilde{I} - a - cI_\mr{bg}$ at every pixel and drop from the set any pixel whose error is more than $2.5\times1.4826\times\mathrm{MAD}(r)$ above the median error. The value of 1.4826 scales the MAD to a standard deviation assuming \(r\) follows a Gaussian distribution \cite{rousseeuwAlternativesMedianAbsolute1993}. The value of 2.5 was chosen empirically. We repeat these steps five times. The final background estimate is $a + cI_\mr{bg}$, and the pixel value containing the actual aurora used for training is obtained as $\tilde{I} - (a + cI_\mr{bg})$. \cref{fig:background_removal} displays the result of our background removal procedure.
\section{Real Observations}

The real dataset is a recording of auroras taken on 26 March 2023 from the same camera locations as in the synthetic dataset. Images are pre-processed according to \cite{fukizawaThreedimensionalIonosphericConductivity2023}. Unlike our synthetic datasets, these real observations don't contain images at 844.6 nm, only 427.8 and 557.7 nm. To compare our method against G-ACT, we use these same pre-processed images as input to our training procedure. In particular, camera sensitivities are correspondingly fixed.

We train our neural field on this dataset in a similar manner to our synthetic datasets. Small per-camera pointing offsets and a one-frame delay of the 557.7 nm images are corrected during training (see the notebook). We use the same variance parameters for the loss, ($\alpha=4$, $\beta=40^2$ R$^2$). A penalty on the linear roughness of the spectra is included with a weight \(\mu_\mr{lin} = 10^5\). We also increase the number of training iterations to 20,000. The training time was 6 minutes.
Re-rendered images from our reconstruction and G-ACT's reconstruction are shown in \cref{fig:rerender_4278} and \cref{fig:rerender_5577}. Note that all views, including these re-renders, were used for optimization. Since the dataset only contains three viewpoints, keeping one held-out view for validation would significantly deteriorate the reconstruction.

\begin{figure}[H]
    \centering
    \includegraphics[width=1\linewidth]{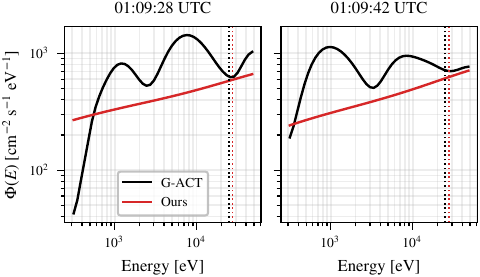}
    \caption{Energy spectra reconstructed by G-ACT and our neural field at the peak total energy flux on the real dataset.}
    \label{fig:real_spectra}
\end{figure}

This large linear-space weight was necessary to obtain reconstructions similar to G-ACT because the log-domain penalty alone is scale-invariant: a spectrum can develop an extreme, physically implausible dynamic range while still having small log-curvature. Adding the linear-domain term prevents this behavior but also smooths out possible spectral structure, producing a nearly monotonic curve instead of the double-humped shape seen in G-ACT's reconstructed spectra as shown in \cref{fig:real_spectra}. G-ACT's spectra aren't, however, a ground-truth we can quantitatively compare against and the comparison should be treated as qualitative.

\end{document}